\documentclass[conference,letterpaper]{IEEEtran}
\IEEEoverridecommandlockouts
\usepackage{cite}
\usepackage{amsmath,amssymb,amsfonts}
\usepackage{algorithmic}
\usepackage{graphicx}
\usepackage{textcomp}
\usepackage{xcolor}

\long\def\invis#1{}  % Used to hide a block of text
\newcommand\gderror[1]{
  \typeout{--------------------------------------------------------------------}
  \typeout{------- #1 ---------}
  \typeout{--------------------------------------------------------------------}
  {\bf #1}
}
\newcounter{gdTmp}
\newcounter{gdLastCount}
\newcommand\maxpage[2][Error]{  %  arg1: error message, arg2: max page number,
\ifnum\value{page}>#2
    \gderror{\Large \bf On page {\thepage} we are past page #2 (too long).   #1 }
\else\fi
}
\newcommand\maxpageSinceLast[2][Error]{  %  arg1: error message, arg2: max page number,
\ifnum \numexpr \value{page} - \value{gdLastCount}\relax>#2
    \gderror{Exceeds max length #2 pages. Page \thepage: #1}
\thepage\else\fi
\setcounter{gdLastCount}{\value{page}}
}
\begin{document}

\title{Self-Supervised Transformer Architecture for Change Detection in Radio Access Networks\\
% {\footnotesize \textsuperscript{*}Note: Sub-titles are not captured in Xplore and should not be used}
% \thanks{Identify applicable funding agency here. If none, delete this.}
}

% \author{\IEEEauthorblockN{1\textsuperscript{st} Igor Kozlov}
% \IEEEauthorblockA{\textit{Samsung AI-Center Montreal} \\
% \textit{Samsung Electronics}\\
% Montreal, Canada \\
% i.kozlov@samsung.com}
% \and
% \IEEEauthorblockN{2\textsuperscript{nd} Dmitriy Rivkin}
% \IEEEauthorblockA{\textit{Samsung AI-Center Montreal} \\
% \textit{Samsung Electronics}\\
% Montreal, Canada \\
% d.rivkin@samsung.com}
% \and
% \IEEEauthorblockN{3\textsuperscript{rd} Wei-Di Chang}
% \IEEEauthorblockA{\textit{Samsung AI-Center Montreal} \\
% \textit{Samsung Electronics}\\
% Montreal, Canada \\
% w.chang1@partner.samsung.com}
% \and
% \IEEEauthorblockN{4\textsuperscript{th} Di Wu}
% \IEEEauthorblockA{\phantom{test}\textit{Samsung AI-Center Montreal}\phantom{test}\\
% \textit{Samsung Electronics}\\
% Montreal, Canada \\
% di.wu1@samsung.com}
% \and
% \IEEEauthorblockN{5\textsuperscript{th} Xue (Steve) Liu}
% \IEEEauthorblockA{\phantom{test}\textit{Samsung AI-Center Montreal}\phantom{test}\\
% \textit{Samsung Electronics}\\
% Montreal, Canada \\
% steve.liu@samsung.com}
% \and
% \IEEEauthorblockN{6\textsuperscript{th} Gregory Dudek}
% \IEEEauthorblockA{\phantom{test}\textit{Samsung AI-Center Montreal}\phantom{test}\\
% \textit{Samsung Electronics}\\
% Montreal, Canada \\
% greg.dudek@samsung.com}
% }

\author{\IEEEauthorblockN{Igor Kozlov\IEEEauthorrefmark{1}, Dmitriy Rivkin\IEEEauthorrefmark{1}, Wei-Di Chang\IEEEauthorrefmark{2},
Di Wu\IEEEauthorrefmark{1}, Xue Liu\IEEEauthorrefmark{1}, Gregory Dudek\IEEEauthorrefmark{1}}% <-this % stops a space
\IEEEauthorblockA{\IEEEauthorrefmark{1}\{i.kozlov, d.rivkin, di.wu1, steve.liu, greg.dudek\}@samsung.com}%
\IEEEauthorblockA{\IEEEauthorrefmark{2}w.chang1@partner.samsung.com}%
}

\maketitle

\begin{abstract}
% ICC track
% Symposium on Selected Areas in Communications:
% Machine Learning for Communications and
% Networking Track
% • Machine learning driven techniques for radio environment awareness and decision making
% • Machine learning for resource management & optimization
% • Unsupervised, semi-supervised, and self-supervised learning approaches to communications
Radio Access Networks (RANs) for telecommunications represent large agglomerations of interconnected hardware consisting of hundreds of thousands of transmitting devices (cells). 
Such networks undergo frequent and often heterogeneous changes caused by network operators, who are seeking to tune their system parameters for optimal performance.
The effects of such changes are challenging to predict and will become even more so with the adoption of 5G/6G networks. 
Therefore, RAN monitoring is vital for network operators. 
We propose a self-supervised learning framework that leverages self-attention and self-distillation for this task.
It works by detecting changes in Performance Measurement data, a collection of time-varying metrics which reflect a set of diverse measurements of the network performance at the cell level.
Experimental results show that our approach outperforms the state of the art by 4\% on a real-world based dataset consisting of about hundred thousands timeseries. 
It also has the merits of being scalable and generalizable. 
This allows it to provide deep insight into the specifics of mode of operation changes while relying minimally on expert knowledge.
\end{abstract}


\section{Introduction}
\label{sec_intro}
This work presents a new change detection system for Performance Measurement (PM) data. The PM system collects thousands of metrics from each cell, quantifying different aspects of the cell's performance, including voice quality indicators, throughput, latency, radio signal strength, interference, etc. The volume of this data is so large that it is not currently feasible for network operators to monitor all of it. As a result, it often happens that when they make  changes to the network configuration, they create a negative impact on some subsystems without realizing it. Alternately, the configuration change may actually have some positive impact that the operators are not aware of, leading them to erroneously roll back the configuration change. Our goal is to develop a system which will monitor PM data and detect all impacts of configuration changes on the system. 

In this work, we address this task by adapting a modern self-supervised representation learning method \cite{dino}, originally developed to deal with large, unlabelled image datasets, to the problem of detecting impacts of configuration changes on PM data. We focus on change, as opposed to anomaly detection. The difference is that change detection focuses on stable changes in operating characteristics, while anomaly detection usually strives to detect outliers, or individual highly unusual data points. This distinction is of great importance to network operators, as transient unusual behaviors are, by definition, self-correcting and therefore do not require any corrective action on the operator's part. On the other hand, when a stable change is detected, it means the cell has begun to operate in a new way, and corrective action may be required.

In order to meet the practical application requirements, the monitoring solution should be:

\begin{itemize}
    \item Scalable: In order to obtain complete monitoring coverage over all of the subsystems reported on by PM data, it is critical to detect changes in a large number of metrics. Therefore, the computational complexity of the change detector should scale linearly with the number of metrics.
    \item Generalizable: RANs usually include hardware from multiple vendors, which also experiences constant software upgrades that modifies the list of recorded PM metrics. The solution should be able to adapt to these changes with minimal expert intervention. 
\end{itemize}

To address these requirements, we monitor individual metrics, as opposed to looking at pair correlations of metrics, which experience polynomial computing resource demand growth \cite{cellpad}. 
This way we can monitor all (usually thousands of) metrics that are being recorded at cell level, which enables deeper insight into the diagnostics of the problem related to the detected change. 
We also point out that when previously co-evolving metrics experience decorrelation, at least one of them experiences a change and thus we expect our method to detect it.
Regarding generalizability requirements, previous approaches (see Section \ref{sec_ad_related_work}) operate over a small set of expert selected metrics, which can vary across vendors and software updates. 
Our solution looks at {\it all} the recorded PM metrics and thus is resilient to such modifications and does not require expert intervention.  

Due to a lack of publicly available datasets for this task, we developed our own. This dataset is based on a proprietary, system-level RAN simulator originally developed and validated to study network parameter optimization \cite{slsstart,slsend}. 

To summarize, the main contributions of this paper are as follows:
\begin{itemize}
    \item We bring attention to the problem of monitoring the network for behavior changes associated with frequent (order of a week) heterogeneous system parameter changes at cell level.
    \item We suggest a scalable and generalizable self-supervised learning framework that leverages self-attention and self-distillation for metric-level change detection.
    \item To evaluate performance on this task, we develop a dataset grounded in real RAN historical observations.
    \item We show superior detection capabilities of our method compared to the current state of the art across non-ML and ML approaches.
\end{itemize}

The rest of the paper is organized as follows: in Section \ref{related_work} we describe previous work related to detecting changes in cellular network metrics. In Section \ref{sec_cpds} we describe in detail the three change point detectors compared in this paper, including our own proposed method. In Section \ref{sec_dataset} we describe the process used for generating the dataset. In Section \ref{sec_comparison} we present a comparison of the performance of the methods introduced in Section \ref{sec_cpds} on the dataset from Section \ref{sec_dataset}. In Section \ref{sec_conclusion} we summarize the work and provide concluding remarks.

\section{Related Work}
\label{related_work}

In this section, we first discuss work on anomaly detection in cellular networks, then review advances in existing approaches for change detection.

\subsection{Anomaly Detection in Cellular Networks}
\label{sec_ad_related_work}

Approaches for anomaly detection in cellular networks vary widely in scope, ranging from detecting anomalies observable in individual KPIs \cite{barreto20053G, qin2018sqoe}, to the multivariate case using multiple KPIs \cite{ciocarlie2013}, or detecting network-level anomalies, taking into account the interconnected structure of the network \cite{chaparro2015detecting, yang2022dynamic}. 
In this paper we focus on the first one. 

Anomalies at the KPI level manifest themselves as irregularities in the signal. 
They can be of short (pointwise) or long lasting (e.g., pattern change in signal's periodicity or amplitude) nature \cite{lai2021revisiting}. 
Often the causes of these types of behavior are different, which motivates the introduction of a special taxonomy (``change detection''), especially as the latter can be more harmful to the network as it indicates significant shift. 
The work we found on anomaly detection in cellular networks does not explicitly make such a distinction. 
We improve on it and also extend to RAN applications the advances of the Change Point Detection field of research, which we briefly review in the next sub-section. 

Approaches for anomaly detection applied to cellular networks span across a wide range: clustering algorithms, such as Self-Organizing Maps \cite{barreto20053G}; ensembling of classification and regression based methods, such as SVM and ARIMA \cite{ciocarlie2013}; correlation and functional analysis \cite{munoz2016}; thresholding and range-based rules \cite{qin2018sqoe}; generative adversarial networks \cite{ad-gan-lstm}; and convolutional (CNNs) and recurrent neural networks, such as LSTMs \cite{rnncnn, ad-lstm}.
\cite{ad-cell-survey} presents a more in depth survey of the existing work in anomaly detection for cellular networks. 
To the best of our knowledge, our approach is the first to make use of the transformer self-attention architecture \cite{transformer} for cellular network anomaly detection.

Most of the work on anomaly detection in RANs uses supervised learning. 
This requires tedious manual labeling of the datasets and also specifying the types of anomaly classes \cite{cellpad, rnncnn}. 
Some attempts to address the labeling challenge by using semi-supervised learning can be found in \cite{ad-semi-supervised-matlab}. 
To the best of our knowledge, we are the first to address both challenges by using self-supervised learning \cite{ssl-vision}, which requires no labels and no prior knowledge on the possible types of anomalies present in the data. 

\subsection{Change Detection}

There is a large body of work on the theoretical and practical aspects of change detection. 
The problem of changepoint detection as the segmentation of the timeseries into segments delimited by the changepoints, thus finding the in-between regions of the changes rather than the points of change themselves, by using the Matrix Profile of the timeseries to detect pattern consistency is formulated in FLUSS and its variant FLOSS \cite{gharghabi2017matrix}. Probabilistic approaches such as BOCPD \cite{bocpd} formulate the problem as estimating a probability distribution over the amount of time since the last changepoint, known as the ``run length". Non-parametric methods such as \cite{matteson2014nonparametric} are based on statistical two-sample testing, samples are deemed anomalous if the test statistic exceeds a specific threshold. KL-CPD \cite{chang2019kernel} extends the kernel changepoint detection non-parametric approach introduced in \cite{harchaoui2008kernel} to allow for deep kernels learnt from data. 
A common underlying paradigm in many learning and non-learning based approaches slides consecutive non-overlapping half-windows over the timeseries and quantifies the mismatch between the left half-window and the right half-window, as they should most differ when the point between the two half-windows is a changepoint. This paradigm is particularly common among unsupervised deep learning approaches, e.g. \cite{betsi}.
One of the most recent instantiations of this approach is TIRE \cite{de2021change}, which we use as a ML baseline, see Section \ref{subsection_TIRE}.

In a recent survey and benchmarking of changepoint detection methods \cite{nonmlcpdsreview}, the authors observe that: first, there is no single Change Point Detector (CPD), which performs best on all datasets, which highlights domain specifics; second, such classical non-ML CPD as Binseg \cite{binseg} performs consistently better across a wide range of datasets compared even to some ML methods, which is the reason we choose as a non-ML baseline, see Section \ref{subsection_binseg}.

\section{Change detectors}
\label{sec_cpds}

In this section we introduce in more details the non-ML (Binseg) and ML (TIRE) baselines, as well as our method (TREX-DINO). 

\subsection{Binseg}
\label{subsection_binseg}

Binseg \cite{binseg} is a sequential algorithm, which consecutively splits timeseries into two windows and calculates the gain obtained from the introduction of this separation by evaluating the difference of the cost function calculated on the original window and the two derived ones. 
This allows to associate specific gain with every (potential change) point of the timeseries. 
We later use these values as a threshold of the Binseg CPD algorithm in the binary classification problem, see Section \ref{sec_comparison} for more details. 

\subsection{TIRE}
\label{subsection_TIRE}

TIRE \cite{de2021change} uses a pair of autoencoders, one in the time domain and one in the frequency domain over consecutive windows of the timeseries to detect changepoints. In order to do so, TIRE learns representations that combine a reconstruction objective and a time-invariant, similarity minimization objective between successive windows. Once the representations are learned, they are used in a window based CPD. Specifically, two sliding windows (backward and forward looking) are used to decide whether the point in the middle is a change point. The similarity measure between data in the sliding windows is computed by applying the cosine distance between their respective projections into the representation space:
\begin{equation}
d(x, y) = 1 - \frac{\sum_{k=1}^{K}{g(x)^{(k)} g(y)^{(k)}}} {\sqrt{\sum_{k=1}^{K}{{g(x)^{(k)}}^{2}}} \sqrt{\sum_{k=1}^{K}{{g(y)^{(k)}}^{2}}}},
\label{eq:distance}
\end{equation}
where $x$ and $y$ are datapoints in the corresponding windows; $g$ is the learned representation, and $K$ is the dimensionality of the representation space. Application of this sliding window procedure yields a new timeseries, $d$. In order to extract change points, $d$ is first filtered, then a peak finding procedure is applied to the result.

\subsection{TREX-DINO}

We propose to learn representations from large, unlabelled datasets using self supervision and transformers, a learning paradigm which has recently enabled impressive performance in natural language processing \cite{bert} and computer vision \cite{vit}. Our method, dubbed TREX-DINO (Timeseries REpresentation eXtraction using DIstillation with NO labels), is an extension of DINO \cite{dino}, a method originally proposed for self-supervised representation learning on images. At a high level, the self-supervised learning objective encourages the model to project multiple random augmentations of the same timeseries into similar representations, while augmentations of different timeseries are encouraged to have different representations. This is achieved using a knowledge distillation \cite{distillation} style student-teacher learning modality. However, instead of having a fixed teacher, we periodically update the teacher weights using an exponential moving average of the student weights. This effectively implements a form of model ensembling in the teacher, akin to Polyak-Ruppert averaging ~\cite{polyak1992acceleration, ruppert1988efficient}.

The student network is trained to match the teacher's distribution over the dimensions of the learned representation using a cross entropy loss. Given a network $g$, input timeseries $x$, the size of the representation $K$, and a temperature parameter $\tau$ which controls the sharpness of the distribution, the probability $P$ over the dimensions of the representation is given by:
\begin{equation}
    P(x)^{(i)} = \frac{exp(g(x)^{(i)} / \tau)}{\sum_{k=1}^{K} exp(g(x)^{(k)}/\tau)}.
\end{equation}

In the standard knowledge distillation setup, the optimization objective would be given by
\begin{equation}
    \min_{\theta_s} H(P_t(x), P_s(x)),
\end{equation}
where $\theta_s$ are the parameters of the student network and $H$ is the cross-entropy loss. However, in order to adapt this to the self-supervised setting, DINO first defines the set $V$, which consists of different augmentations of the same timeseries. Critically, two of the items in $V$, $x_1^g$ and $x_2^g$, are so called ``global'' views. These are large crops of the  original timeseries, which include most of it. The rest of the augmentations of $x$ in $V$ feature ``local'' crops, which remove a larger fraction of the original timeseries. This style of augmentation is known in the computer vision literature as ``multi-crop" \cite{multi-crop}. The teacher is then used to compute representations for the global views only, while the student computes representations for all of them. In other words, we aim to optimize the following optimization objective:
\begin{equation}
    \min_{\theta_s} \sum_{x \in \{x_1^g, x_2^g\}} \sum_{x' \in V, x` \neq x} H(P_t(x), P_s(x')).
\end{equation}

The weights of the teacher, $\theta_t$, are updated from those of the student using the following update rule ($\lambda$ is a hyperparameter):
\begin{equation}
    \theta_t = \lambda\theta_t + (1-\lambda)\theta_s.
\end{equation}

Collapse (i.e. the network learning to embed all of the timeseries into exactly the same representation) is avoided by centering the outputs of the teacher network, $g_t(x) = g_t(x) + c$, where the center, $c$, is dynamically adjusted as follows ($m$ is a hyperparameter, $B$ is the batch size):

\begin{equation}
    c = mc + (1-m) \frac{1}{B} \sum_{i=1}^{B}g_t(x_i).
\end{equation}
See \cite{dino} for a more complete treatment of the approach.

DINO was originally targeted at images and implemented using a vision transformer (ViT) \cite{vit}. In order to adapt this method to the timeseries case, some modifications were needed. First, we adapted the ViT to accept timeseries as input. The ViT architecture splits images into patches, each of which are passed through a convolutional layer and flattened to create a series of tokens which are used as input to the transformer. TREX-DINO adapts these from 2-dimensional convolutions, appropriate for images, to 1-dimensional ones better suited to timeseries. We keep the channel dimension, so TREX-DINO supports multi-dimensional timeseries, although in this paper we focus on the single dimensional case. Furthermore, DINO augmentations include color jittering, Gaussian blur, and solarization, as well as multi-crop. TREX-DINO keeps the Gaussian blur and multi-crop, but replaces color jittering and solarization with additive Gaussian noise.

Once the representations are learned, they are used to detect changes using the sliding window method as outlined in Section \ref{subsection_TIRE}.

\section{Dataset}
\label{sec_dataset}

One can observe variance in CPDs performance on timeseries of various origin \cite{nonmlcpdsreview}, which is due to data peculiarities pertinent to each domain (e.g., seasonality, recording periodicity and length, change type and their occurrence frequency).
This is the main reason that domain specific task data is of primary importance for evaluating the performance of proposed methods.

\begin{figure}[htbp]
  \begin{center}
    \includegraphics[width=0.7\columnwidth]{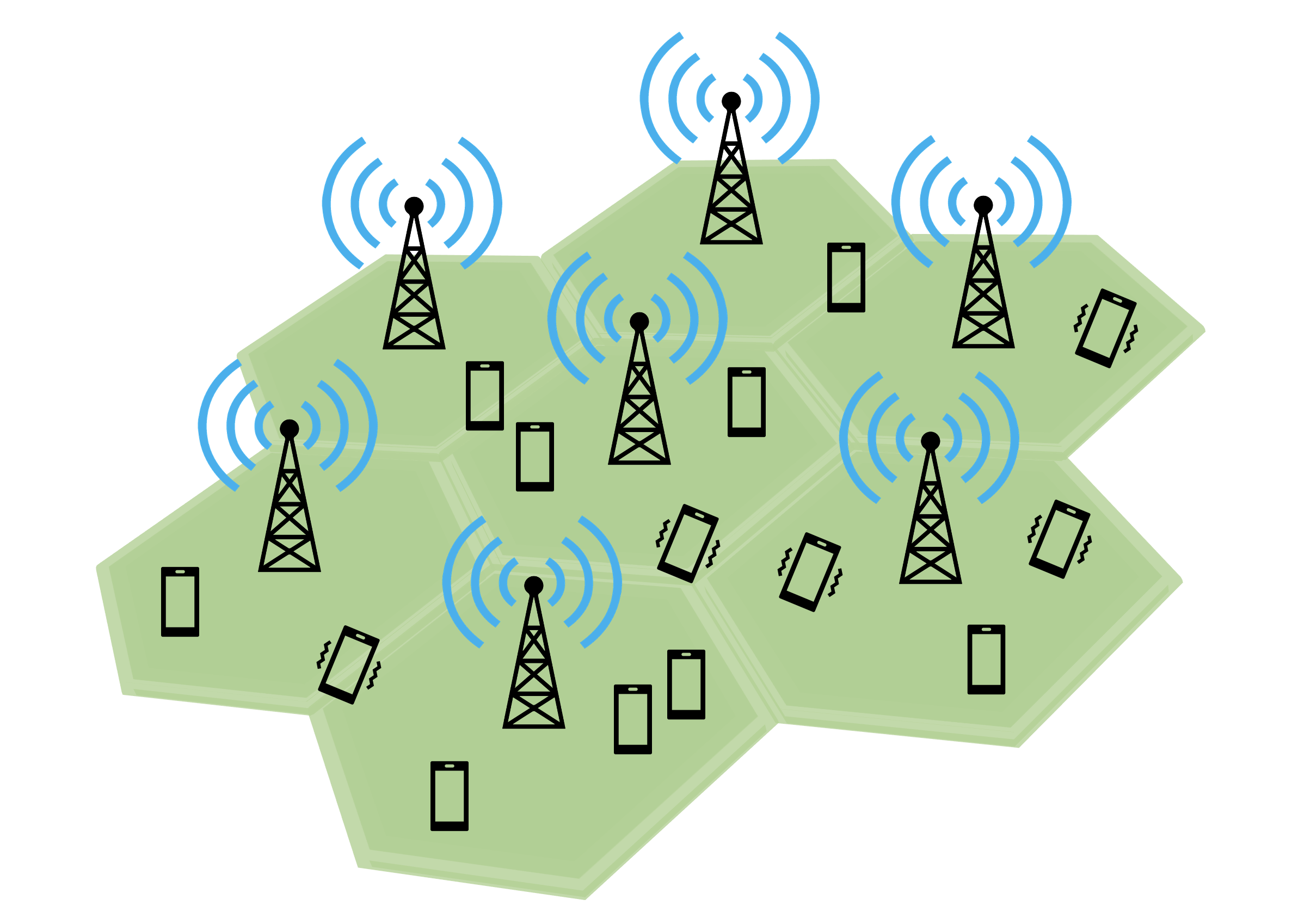}
  \end{center}
\caption{A schematic illustration of an SLS environment that consists of 7 sites, which are serving a number of active and idle User Equipments (UEs).}
\label{fig_hex7}
\end{figure}

In this paper we used a proprietary System Level Simulator (SLS) \cite{slsstart,slsend}, which was specifically developed for studying effects of system parameter changes on cellular network performance, and is based on real-world RAN data.
It models traffic in RANs for specific site configuration architectures. 
We utilize the one schematically depicted in Fig. \ref{fig_hex7}. 
Every site operates over 4 frequency bands in each of 3 non-overlapping sectors. 
UEs are initially uniformly distributed across all sites and randomly walk with constant velocity, or stay in one place. 
They contribute to the traffic load of the network by downloading files in a stochastic manner. 
This process is modelled to reproduce historically observed RAN data scenarios, with UE behavior parameters varying accordingly throughout the day \cite{slsstart,slsend}. 
Cells transmitting data to UEs also have realistic system parameters preset, but these do not automatically vary with time. 

The main focus of this paper is to detect changes in network operations by monitoring PM data. 
Each of the cells records 28 various timeseries. 
They include Key Performance Indicators (KPIs) closely monitored by network operators, e.g. down-link throughput, as well as metrics commonly used in literature for modeling of RAN performance, e.g. physical resource block usage and number of active UEs. 
All aforementioned timeseries are hand picked by experts and are often transformed using domain knowledge, e.g. throughput is a ratio of the radio link data and delay. 
As discussed in Section \ref{sec_intro}, we seek a generalizable approach to RAN monitoring, and thus we also include in our analysis an extensive list of low level subsystem metrics, e.g. the number of successful handovers, 
treating them all on par. 

In addition to the aforementioned variability coming from historical RAN data UE parameters, our dataset includes two additional degrees of freedom. 
First, infrequent system parameter changes. 
Those affect SLS parameters describing cell subsystems. 
This is done to model RAN performance improvement/degradation due to endogenous, e.g. Configuration Management (CM) changes, and exogenous, e.g. hardware failures, causes that represent interest to the network operators. 
Second, pseudo-random changes. 
These affect UE dynamics and are useful for verifying stability of the network operations and enriching the dataset with realistic variance. 

The main difference between our dataset and previous applications of SLS are as follows:
\begin{itemize}
\item We are not interested in the dynamics of individual UEs it can provide, because for privacy and technical reasons it is easier to get access to aggregated over UEs PM data to base our change detection on.
\item For accurately modeling diverse traffic scenarios, SLS uses multiple historical network observations from field deployments, which have finite (maximum week-long) recording length. ML and especially deep learning methods that we propose to use (transformers) extract features without human engineering effort. In order to do it efficiently they need to have access to a large corpus of data, e.g. to have enough historical information to detect lasting operating modes and changes between them. For that reason, we produce samples of our dataset by concatenating week-long outputs of SLS data.
\end{itemize}

Here is the detailed procedure that we use to construct a sample of the dataset. 
We run week-long SLS simulations for the network described in Fig. \ref{fig_hex7} with combinations of 5 different system parameter sets and 10 various seeds. 
We then choose a pair of distinct system parameter changes, $\kappa$, and seeds, $\rho$, to concatenate week-long timeseries, $\tilde x$, into four week-long samples, $\hat x$:
\begin{equation}
  \hat x^m_{t}(\rho_{r_1}, \kappa_{k_1}) = 
  \begin{cases}
    \tilde x^m_{t}(\rho_{r_1}, \kappa_{k_1})& \text{if} \, t \in [0, 168),\\
    \tilde x^m_{t-168}(\rho_{r_2}, \kappa_{k_1})& \text{if} \, t \in [168, 336),\\
    \tilde x^m_{t-336}(\rho_{r_1}, \kappa_{k_2})& \text{if} \, t \in [336, 504),\\
    \tilde x^m_{t-504}(\rho_{r_2}, \kappa_{k_2})& \text{if} \, t \in [504, 672),
  \end{cases}\,
\end{equation}
where $\tilde x^m_t(\rho_r, \kappa_k)$ represents time $t$ (in hours) metric $m$ recording of the week-long simulation ran with seed $r$ and system parameters set $k$, and subscripts indicate whether subscribed variables have equal values.

\begin{figure}[htbp]
  \begin{center}    
    \includegraphics[width=0.8\columnwidth]{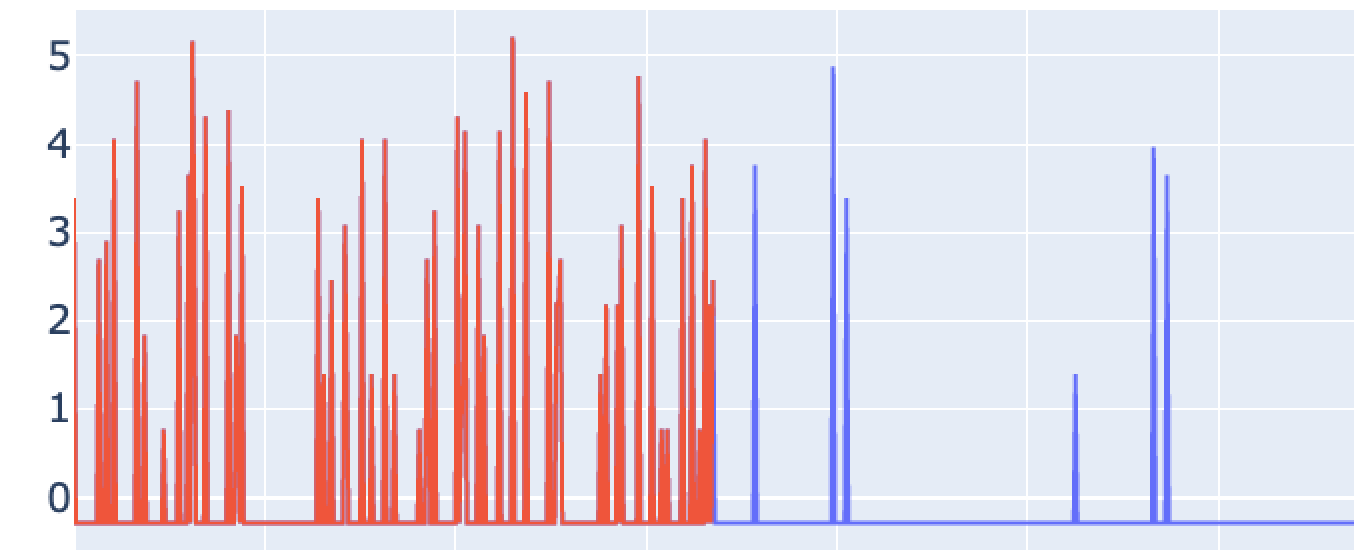}
    \includegraphics[width=0.8\columnwidth]{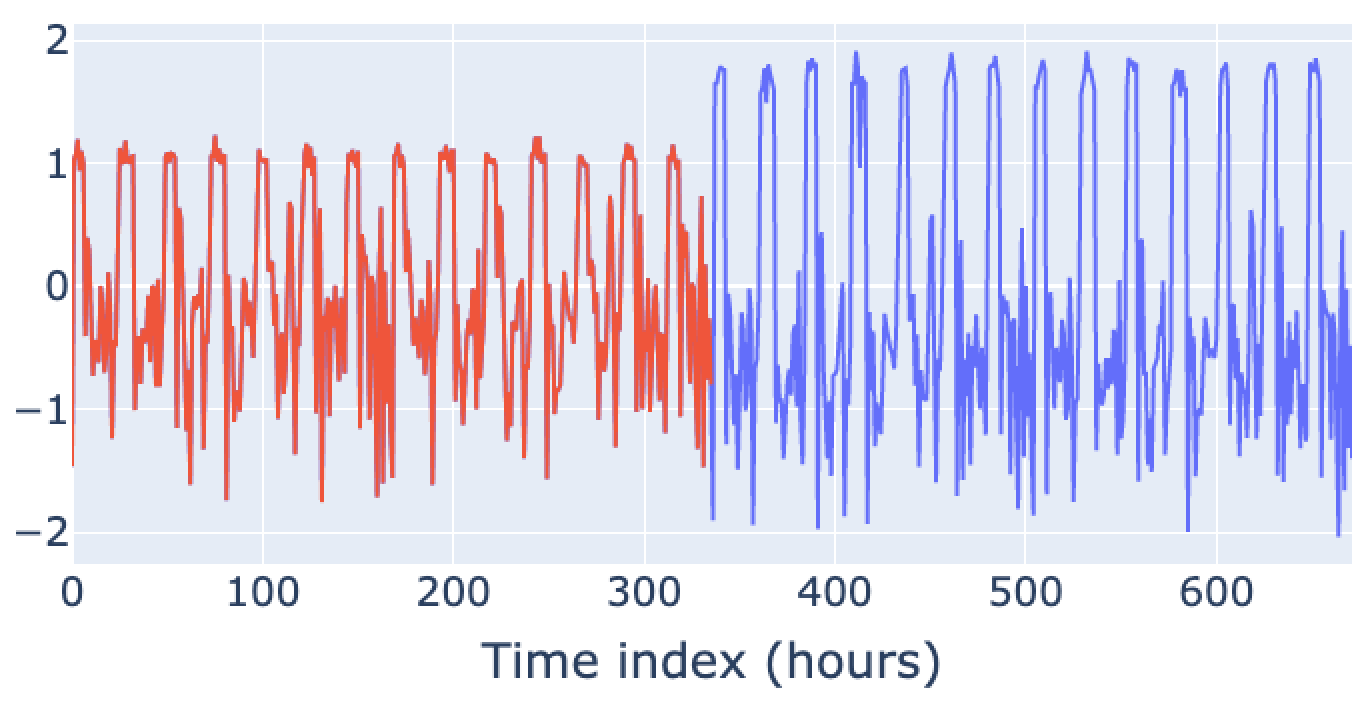}
  \end{center}
\caption{Examples of discrete (top) and continuous (bottom) timeseries data created using SLS. The color coding indicates distinct system parameters sets.}
\label{fig_timeseries}
\end{figure}

Finally, we preprocess the constructed timeseries, $\hat x$, to smooth the data and make it suitable for use in neural networks by the standard procedure of using logarithmic transformation, removing the mean, and scaling to unit variance \cite{scikit-learn, paszke2017automatic}, to obtain our dataset, $x$. 
One can see examples of the resulting timeseries, $x$, in Fig. \ref{fig_timeseries}.
Timeseries pattern changes like the ones in Fig. \ref{fig_timeseries} have no ``drop'' \cite{cellpad}, but clearly indicate a shift in operating characteristics caused by a modification of a system parameter that network operators are interested in learning about. 
We include them to highlight new types of domain specific anomalies that are of particular relevance in telecom. 

To create a test set of labels to evaluate CPDs' performance and understanding that system parameters switch might not affect every metric, we manually verified timeseries by leveraging the knowledge of the dataset construction (location of the change point and week-long nature of the subsegments), by clustering samples using week-over-week statistics, and keeping for validation samples only from those clusters that clearly had a present or absent mode of operations change, which resulted in 78455 samples without pattern changes and 14301 samples with it.  

\section{Performance Evaluation}
\label{sec_comparison}
In this section, we evaluate performance of CPDs described in Section \ref{sec_cpds} using the dataset presented in Section \ref{sec_dataset}. In order to simplify the comparison, we evaluate performance on a binary classification task where competing algorithms must identify whether a changepoint is present in the timeseries.

We use the Binseg implementation by Truong et al. from \cite{ruptures} with RBF cost function \cite{rbf-cost}. 

TIRE and TREX-DINO were both trained on a randomly selected half of the dataset and tested all together with Binseg on the other half. 

Hyperparameters' values used for training TIRE and TREX-DINO were based on \cite{de2021change,dino}.  
Specifically, for TIRE, we used 3 parallel autoencoders in both timeseries and frequency domains; 
for the DINO subpart of TREX-DINO, we used 12 self-attention heads with embedding dimensions size of 768, Gaussian noise with standard deviation of 0.3, and Gaussian blur with kernel standard deviation uniformly distributed in the \mbox{[0.1, 2.]} range and applied on average in half of the augmentation calls. 

The only parameter that we set manually based on the domain specifics was the half window size of 168 (equals $24\times7$). 
In effect, this biases the system to perform week over week comparisons, which is a common practice among telecommunications operators in their assessments.

Another important hyperparameter present in all compared models is the detection threshold: timeseries with predicted value above it are considered to have a change point. 
Instead of using the validation set to tune it for every method, we chose to compare methods numerically at their top performance and visually over the entire range of the values: this provides more robustness due to the narrow peak shape of some of the performance curves. 
Due to the different nature of the algorithms and thus resulting threshold values, we plot curves after uniform normalization to the $[0, 1]$ range.

\begin{figure}[htbp]
\begin{center}
\includegraphics[width=0.9\columnwidth]{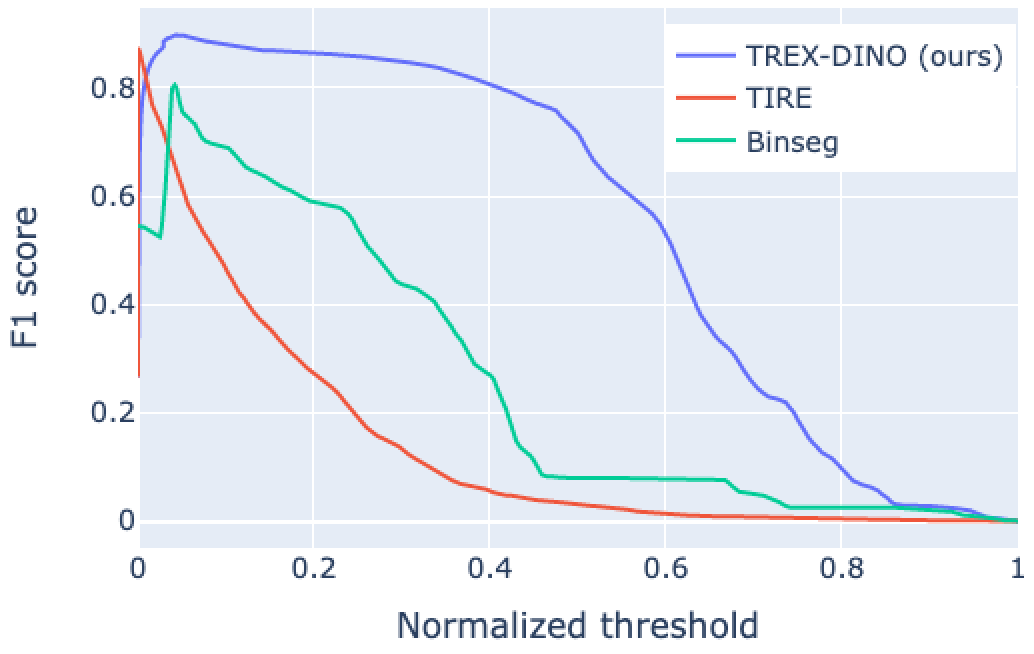}    
\end{center}
\caption{Differential comparison of CPDs' performance as a function of normalized detection threshold.}
\label{fig_f1}
\end{figure}

The F1 curves for the three methods are shown in Figure \ref{fig_f1}. The maximum F1 scores  and the area under precision-recall curve (PR AUC) attained by each method are shown in Table \ref{tab_f1}. In addition to achieving the highest max F1 score of the three, TREX-DINO also attains a high F1 score over a much wider range of normalized threshold values, indicating that it is less sensitive to the choice of the threshold. This is also reflected in its superior PR AUC score.

\begin{table}[htbp]
\caption{Aggregate comparison of CPDs using  F1 max and PR AUC.}
\begin{center}
\begin{tabular}{|c|c|c|c|}
\hline
\textbf{Metric}&\textbf{Binseg}&\textbf{TIRE}&\textbf{TREX-DINO (ours)}\\
\hline
\textbf{F1 score} & 0.80 & 0.87 {(+9\%)} & {\bf 0.90 {(+13\%)}}\\
\hline
\textbf{PR AUC} & 0.82 & 0.91 {(+11\%)} & {\bf 0.97 {(+18\%)}}\\
\hline
\textbf{Rank}& Baseline & Better & Best\\
\hline
\end{tabular}
\label{tab_f1}
\end{center}
\end{table}

\section{Conclusion}
\label{sec_conclusion}
In this paper, we discussed the importance of monitoring RANs for frequent change detection and proposed a novel change point detector to address this problem. 
In the face of the lack of public benchmarks devoted to this task, we developed a dataset that reproduces the challenges that plague RAN operators on a daily basis. 
Experiments demonstrated the superior performance of our approach compared to existing methods. 
We are confident that our framework, proven in the fields of computer vision and natural language processing and based on a self-supervised transformer architecture with self-distillation, will continue to show good performance characteristics in real world RAN applications.

In order to present changes detected at metric level over hundreds of thousands of RAN devices to human network operators for actionable outcomes, one needs to aggregate them into meaningful events that share the same root cause. 
We leave results of our research effort applied in this direction for future publication. 
However, the foundation of any such aggregation scheme must necessarily be accurate change detection at the metric level, which is the problem that is best addressed by our method, TREX-DINO.

\section*{Acknowledgment}

The authors thank Ekram Hossain, Perouz Taslakian, Amal Feriani, and Bassem Maraach for useful discussions.

% The preferred spelling of the word ``acknowledgment'' in America is without
% an ``e'' after the ``g''. Avoid the stilted expression ``one of us (R. B.
% G.) thanks $\ldots$''. Instead, try ``R. B. G. thanks$\ldots$''. Put sponsor
% acknowledgments in the unnumbered footnote on the first page.

\bibliographystyle{IEEEtran}

\maxpage[Pre-bibliography document length limit exceeded]{6}

\bibliography{bibliography}

% Generated by IEEEtran.bst, version: 1.12 (2007/01/11)
\begin{thebibliography}{10}
\providecommand{\url}[1]{#1}
\csname url@samestyle\endcsname
\providecommand{\newblock}{\relax}
\providecommand{\bibinfo}[2]{#2}
\providecommand{\BIBentrySTDinterwordspacing}{\spaceskip=0pt\relax}
\providecommand{\BIBentryALTinterwordstretchfactor}{4}
\providecommand{\BIBentryALTinterwordspacing}{\spaceskip=\fontdimen2\font plus
\BIBentryALTinterwordstretchfactor\fontdimen3\font minus
  \fontdimen4\font\relax}
\providecommand{\BIBforeignlanguage}[2]{{%
\expandafter\ifx\csname l@#1\endcsname\relax
\typeout{** WARNING: IEEEtran.bst: No hyphenation pattern has been}%
\typeout{** loaded for the language `#1'. Using the pattern for}%
\typeout{** the default language instead.}%
\else
\language=\csname l@#1\endcsname
\fi
#2}}
\providecommand{\BIBdecl}{\relax}
\BIBdecl

\bibitem{dino}
\BIBentryALTinterwordspacing
M.~Caron, H.~Touvron, I.~Misra, H.~Jégou, J.~Mairal, P.~Bojanowski, and
  A.~Joulin, ``Emerging properties in self-supervised vision transformers,''
  2021. [Online]. Available: \url{https://arxiv.org/abs/2104.14294}
\BIBentrySTDinterwordspacing

\bibitem{cellpad}
J.~Wu, P.~P.~C. Lee, Q.~Li, L.~Pan, and J.~Zhang, ``Cellpad: Detecting
  performance anomalies in cellular networks via regression analysis,'' in
  \emph{2018 IFIP Networking Conference (IFIP Networking) and Workshops}, 2018,
  pp. 1--9.

\bibitem{slsstart}
J.~Kang, X.~Chen, D.~Wu, Y.~T. Xu, X.~Liu, G.~Dudek, T.~Lee, and I.~Park,
  ``Hierarchical policy learning for hybrid communication load balancing,'' in
  \emph{ICC 2021 - IEEE International Conference on Communications}, 2021, pp.
  1--6.

\bibitem{slsend}
A.~Feriani, D.~Wu, Y.~T. Xu, J.~Li, S.~Jang, E.~Hossain, X.~Liu, and G.~Dudek,
  ``Multiobjective load balancing for multiband downlink cellular networks: A
  meta- reinforcement learning approach,'' \emph{IEEE Journal on Selected Areas
  in Communications}, vol.~40, no.~9, pp. 2614--2629, 2022.

\bibitem{barreto20053G}
G.~Barreto, J.~Mota, L.~Souza, R.~Frota, and L.~Aguayo, ``Condition monitoring
  of 3g cellular networks through competitive neural models,'' \emph{IEEE
  Transactions on Neural Networks}, vol.~16, no.~5, pp. 1064--1075, 2005.

\bibitem{qin2018sqoe}
X.~Qin, S.~Tang, X.~Chen, D.~Miao, and G.~Wei, ``Sqoe kqis anomaly detection in
  cellular networks: Fast online detection framework with hourglass
  clustering,'' \emph{China Communications}, vol.~15, no.~10, pp. 25--37, 2018.

\bibitem{ciocarlie2013}
G.~F. Ciocarlie, U.~Lindqvist, S.~Nováczki, and H.~Sanneck, ``Detecting
  anomalies in cellular networks using an ensemble method,'' in
  \emph{Proceedings of the 9th International Conference on Network and Service
  Management (CNSM 2013)}, 2013, pp. 171--174.

\bibitem{chaparro2015detecting}
C.~Chaparro and W.~Eberle, ``Detecting anomalies in mobile telecommunication
  networks using a graph based approach,'' in \emph{The Twenty-Eighth
  International Flairs Conference}, 2015.

\bibitem{yang2022dynamic}
C.~Yang, H.~Song, M.~Tang, L.~Danon, and Y.~Vigfusson, ``Dynamic network
  anomaly modeling of cell-phone call detail records for infectious disease
  surveillance,'' in \emph{Proceedings of the 28th ACM SIGKDD Conference on
  Knowledge Discovery and Data Mining}, 2022, pp. 4733--4742.

\bibitem{lai2021revisiting}
K.-H. Lai, D.~Zha, J.~Xu, Y.~Zhao, G.~Wang, and X.~Hu, ``Revisiting time series
  outlier detection: Definitions and benchmarks,'' in \emph{Proceedings of the
  Neural Information Processing Systems Track on Datasets and Benchmarks},
  J.~Vanschoren and S.~Yeung, Eds., vol.~1, 2021.

\bibitem{munoz2016}
P.~Muñoz, R.~Barco, I.~Serrano, and A.~Gómez-Andrades, ``Correlation-based
  time-series analysis for cell degradation detection in son,'' \emph{IEEE
  Communications Letters}, vol.~20, no.~2, pp. 396--399, 2016.

\bibitem{ad-gan-lstm}
J.~Huang, E.~Kurniawan, and S.~Sun, ``Cellular kpi anomaly detection with gan
  and time series decomposition,'' in \emph{ICC 2022 - IEEE International
  Conference on Communications}, 2022, pp. 4074--4079.

\bibitem{rnncnn}
M.~R. Tanhatalab, H.~Yousefi, H.~M. Hosseini, M.~M. Bonab, V.~Fakharian, and
  H.~Abarghouei, ``Deep ran: A scalable data-driven platform to detect
  anomalies in live cellular network using recurrent convolutional neural
  network,'' in \emph{2020 IEEE 18th World Symposium on Applied Machine
  Intelligence and Informatics (SAMI)}, 2020, pp. 269--274.

\bibitem{ad-lstm}
S.~M.~A. Al~Mamun and M.~Beyaz, ``Lstm recurrent neural network (rnn) for
  anomaly detection in cellular mobile networks,'' in \emph{Machine Learning
  for Networking}, {\'E}.~Renault, P.~M{\"u}hlethaler, and S.~Boumerdassi,
  Eds.\hskip 1em plus 0.5em minus 0.4em\relax Cham: Springer International
  Publishing, 2019, pp. 222--237.

\bibitem{ad-cell-survey}
\BIBentryALTinterwordspacing
B.~Li, S.~Zhao, R.~Zhang, Q.~Shi, and K.~Yang, ``Anomaly detection for cellular
  networks using big data analytics,'' \emph{IET Communications}, vol.~13,
  no.~20, pp. 3351--3359, 2019. [Online]. Available:
  \url{https://ietresearch.onlinelibrary.wiley.com/doi/abs/10.1049/iet-com.2019.0765}
\BIBentrySTDinterwordspacing

\bibitem{transformer}
\BIBentryALTinterwordspacing
A.~Vaswani, N.~Shazeer, N.~Parmar, J.~Uszkoreit, L.~Jones, A.~N. Gomez,
  L.~Kaiser, and I.~Polosukhin, ``Attention is all you need,'' 2017. [Online].
  Available: \url{https://arxiv.org/abs/1706.03762}
\BIBentrySTDinterwordspacing

\bibitem{ad-semi-supervised-matlab}
B.~Hussain, Q.~Du, and P.~Ren, ``Semi-supervised learning based big data-driven
  anomaly detection in mobile wireless networks,'' \emph{China Communications},
  vol.~15, no.~4, pp. 41--57, 2018.

\bibitem{ssl-vision}
\BIBentryALTinterwordspacing
J.-B. Grill, F.~Strub, F.~Altché, C.~Tallec, P.~H. Richemond, E.~Buchatskaya,
  C.~Doersch, B.~A. Pires, Z.~D. Guo, M.~G. Azar, B.~Piot, K.~Kavukcuoglu,
  R.~Munos, and M.~Valko, ``Bootstrap your own latent: A new approach to
  self-supervised learning,'' 2020. [Online]. Available:
  \url{https://arxiv.org/abs/2006.07733}
\BIBentrySTDinterwordspacing

\bibitem{gharghabi2017matrix}
S.~Gharghabi, Y.~Ding, C.-C.~M. Yeh, K.~Kamgar, L.~Ulanova, and E.~Keogh,
  ``Matrix profile viii: domain agnostic online semantic segmentation at
  superhuman performance levels,'' in \emph{2017 IEEE international conference
  on data mining (ICDM)}.\hskip 1em plus 0.5em minus 0.4em\relax IEEE, 2017,
  pp. 117--126.

\bibitem{bocpd}
\BIBentryALTinterwordspacing
R.~P. Adams and D.~J.~C. MacKay, ``Bayesian online changepoint detection,''
  2007. [Online]. Available: \url{https://arxiv.org/abs/0710.3742}
\BIBentrySTDinterwordspacing

\bibitem{matteson2014nonparametric}
D.~S. Matteson and N.~A. James, ``A nonparametric approach for multiple change
  point analysis of multivariate data,'' \emph{Journal of the American
  Statistical Association}, vol. 109, no. 505, pp. 334--345, 2014.

\bibitem{chang2019kernel}
\BIBentryALTinterwordspacing
W.~Chang, C.~Li, Y.~Yang, and B.~P{\'{o}}czos, ``Kernel change-point detection
  with auxiliary deep generative models,'' in \emph{7th International
  Conference on Learning Representations, {ICLR} 2019, New Orleans, LA, USA,
  May 6-9, 2019}.\hskip 1em plus 0.5em minus 0.4em\relax OpenReview.net, 2019.
  [Online]. Available: \url{https://openreview.net/forum?id=r1GbfhRqF7}
\BIBentrySTDinterwordspacing

\bibitem{harchaoui2008kernel}
Z.~Harchaoui, E.~Moulines, and F.~Bach, ``Kernel change-point analysis,'' in
  \emph{Advances in Neural Information Processing Systems}, D.~Koller,
  D.~Schuurmans, Y.~Bengio, and L.~Bottou, Eds., vol.~21.\hskip 1em plus 0.5em
  minus 0.4em\relax Curran Associates, Inc., 2008.

\bibitem{betsi}
\BIBentryALTinterwordspacing
W.-H. Lee, J.~Ortiz, B.~Ko, and R.~Lee, ``Time series segmentation through
  automatic feature learning,'' 2018. [Online]. Available:
  \url{https://arxiv.org/abs/1801.05394}
\BIBentrySTDinterwordspacing

\bibitem{de2021change}
T.~De~Ryck, M.~De~Vos, and A.~Bertrand, ``Change point detection in time series
  data using autoencoders with a time-invariant representation,'' \emph{IEEE
  Transactions on Signal Processing}, vol.~69, pp. 3513--3524, 2021.

\bibitem{nonmlcpdsreview}
\BIBentryALTinterwordspacing
G.~J. J. v.~d. Burg and C.~K.~I. Williams, ``An evaluation of change point
  detection algorithms,'' 2020. [Online]. Available:
  \url{https://arxiv.org/abs/2003.06222}
\BIBentrySTDinterwordspacing

\bibitem{binseg}
A.~Scott and M.~Knott, ``A cluster analysis method for grouping means in the
  analysis of variance,'' \emph{Biometrics}, vol.~30, p. 507, 1974.

\bibitem{bert}
\BIBentryALTinterwordspacing
J.~Devlin, M.-W. Chang, K.~Lee, and K.~Toutanova, ``Bert: Pre-training of deep
  bidirectional transformers for language understanding,'' 2018. [Online].
  Available: \url{https://arxiv.org/abs/1810.04805}
\BIBentrySTDinterwordspacing

\bibitem{vit}
\BIBentryALTinterwordspacing
A.~Dosovitskiy, L.~Beyer, A.~Kolesnikov, D.~Weissenborn, X.~Zhai,
  T.~Unterthiner, M.~Dehghani, M.~Minderer, G.~Heigold, S.~Gelly, J.~Uszkoreit,
  and N.~Houlsby, ``An image is worth 16x16 words: Transformers for image
  recognition at scale,'' 2020. [Online]. Available:
  \url{https://arxiv.org/abs/2010.11929}
\BIBentrySTDinterwordspacing

\bibitem{distillation}
\BIBentryALTinterwordspacing
G.~Hinton, O.~Vinyals, and J.~Dean, ``Distilling the knowledge in a neural
  network,'' 2015. [Online]. Available: \url{https://arxiv.org/abs/1503.02531}
\BIBentrySTDinterwordspacing

\bibitem{polyak1992acceleration}
B.~T. Polyak and A.~B. Juditsky, ``Acceleration of stochastic approximation by
  averaging,'' \emph{SIAM journal on control and optimization}, vol.~30, no.~4,
  pp. 838--855, 1992.

\bibitem{ruppert1988efficient}
D.~Ruppert, ``Efficient estimations from a slowly convergent robbins-monro
  process,'' Cornell University Operations Research and Industrial Engineering,
  Tech. Rep., 1988.

\bibitem{multi-crop}
\BIBentryALTinterwordspacing
M.~Caron, I.~Misra, J.~Mairal, P.~Goyal, P.~Bojanowski, and A.~Joulin,
  ``Unsupervised learning of visual features by contrasting cluster
  assignments,'' 2020. [Online]. Available:
  \url{https://arxiv.org/abs/2006.09882}
\BIBentrySTDinterwordspacing

\bibitem{scikit-learn}
F.~Pedregosa, G.~Varoquaux, A.~Gramfort, V.~Michel, B.~Thirion, O.~Grisel,
  M.~Blondel, P.~Prettenhofer, R.~Weiss, V.~Dubourg, J.~Vanderplas, A.~Passos,
  D.~Cournapeau, M.~Brucher, M.~Perrot, and E.~Duchesnay, ``Scikit-learn:
  Machine learning in {P}ython,'' \emph{Journal of Machine Learning Research},
  vol.~12, pp. 2825--2830, 2011.

\bibitem{paszke2017automatic}
A.~Paszke, S.~Gross, S.~Chintala, G.~Chanan, E.~Yang, Z.~DeVito, Z.~Lin,
  A.~Desmaison, L.~Antiga, and A.~Lerer, ``Automatic differentiation in
  pytorch,'' 2017.

\bibitem{ruptures}
\BIBentryALTinterwordspacing
C.~Truong, L.~Oudre, and N.~Vayatis, ``A review of change point detection
  methods,'' \emph{CoRR}, vol. abs/1801.00718, 2018. [Online]. Available:
  \url{http://arxiv.org/abs/1801.00718}
\BIBentrySTDinterwordspacing

\bibitem{rbf-cost}
\BIBentryALTinterwordspacing
S.~Arlot, A.~Celisse, and Z.~Harchaoui, ``A kernel multiple change-point
  algorithm via model selection,'' 2012. [Online]. Available:
  \url{https://arxiv.org/abs/1202.3878}
\BIBentrySTDinterwordspacing

\end{thebibliography}

\end{document}